\documentclass{6838publ}
\usepackage{6838}

\SpecialIssuePaper
\electronicVersion 
\ifpdf \usepackage[pdftex]{graphicx} \pdfcompresslevel=9
\else \usepackage[dvips]{graphicx} \fi

\PrintedOrElectronic

\usepackage{t1enc,dfadobe}
\usepackage{egweblnk}
\usepackage{cite}
\usepackage{lipsum}
\usepackage{amsmath}
\usepackage{amsfonts}

\usepackage{algorithm}
\usepackage{algorithmic}

\graphicspath{ {images/} }

\title[Instance Segmentation for Point Sets]{Instance Segmentation for Point Sets}

\author[A.\ Talwar \& J.\ Laasri]
       {Abhimanyu Talwar$^1$
        and Julien Laasri$^{1}$
        \\
         $^1$Harvard John A. Paulson School of Engineering \\  and Applied Sciences
       }

\begin{document}

\teaser{
 \includegraphics[width=.75\linewidth]{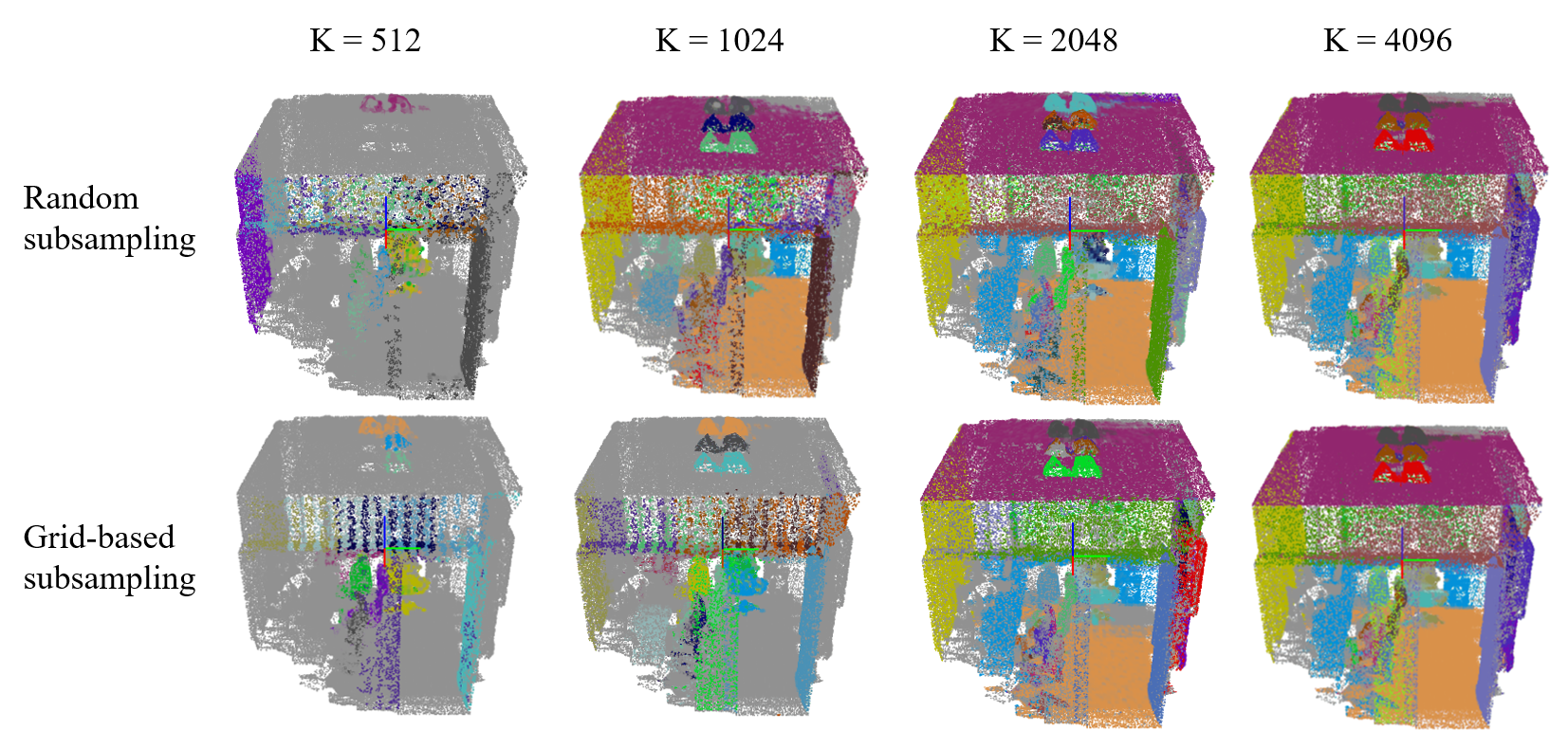}
 \centering
  \caption{Instance segmentation for different numbers of randomly or grid-based selected landmarks. $K$ is the number of sub-sampled landmark points.}
\label{fig:teaser}
}

\maketitle

\begin{abstract}
Recently proposed neural network architectures like PointNet \cite{origpointnet} and PointNet++ \cite{pointnet-plus} have made it possible to apply Deep Learning to 3D point sets. The feature representations of shapes learned by these two networks enabled training classifiers for Semantic Segmentation, and more recently for Instance Segmentation via the Similarity Group Proposal Network (SGPN) \cite{sgpn}. One area of improvement which has been highlighted by SGPN's authors, pertains to use of memory intensive similarity matrices which occupy memory quadratic in the number of points. In this report, we attempt to tackle this issue through use of two sampling based methods, which compute Instance Segmentation on a sub-sampled Point Set, and then extrapolate labels to the complete set using the nearest neigbhour approach. While both approaches perform equally well on large sub-samples, the random-based strategy gives the most improvements in terms of speed and memory usage.
\end{abstract}

\section{Introduction}

Application of Deep Learning based methods to Point Sets is challenging because we want a neural network to be invariant to any permutation of the Point Set. An image or a voxel representation has a sense of order, i.e. the pixels or voxels at adjacent indices are also adjacent in location in the physical space. However this is not true for Point Sets, which are unordered in nature. Qi Charles et al., proposed the PointNet \cite{origpointnet} and PointNet++ \cite{pointnet-plus} architectures which resolved this issue through use of a symmetric function, the $\textbf{MAXPOOLING}$ layer, to aggregate pointwise features into a global shape descriptor. The authors trained Semantic Segmentation classifiers on top of these networks and reported state-of-the-art performance on a few benchmarks.

The PointNet network architectures were successfully used by Wang et al. \cite{sgpn} to do Instance Segmentation. However one of the areas for improvement noted by the authors was reduction in the memory requirement. Specifically, SGPN relies on computation of an $N \times N$ similarity matrix, where $N$ is the number of points in the Point Set.

In this report we address the issue of memory intensiveness through use of sampling and interpolation. Specifically, instead of using an entire point set, we sub-sample it using two approaches which we refer to as (1) Random, and (2) Grid-based. We then compute Instance Segmentation labels for this sub-sampled Point Set, and then extrapolate the labels to the complete Point Set through use of a nearest neighbour (in the Euclidean space) approach. Using these two approaches, we were able to obtain performance (measured using the Mean Average Precision score) similar to that using the full Point Set, albeit in a shorter time and with a reduced memory requirement.

\section{Related Work}


Deep Learning based approaches for point set data include Voxelnet \cite{voxnet} which entails dividing a point set into equally space voxels, and using 3D convolution based model for representation learning. This approach is directly transcribed from the Computer Vision world where 2D Convolutions obtained state-of-the-art results in many applications including image classification \cite{cnn}, style transfer \cite{CycleGAN2017}, depth inference \cite{megadepth}, scene segmentation \cite{segmentation}.

 While these convolutions generalize well to point sets, one of the main drawback of this approach is that it makes the data unnecessary voluminous. The original PointNet architecture proposed by \cite{origpointnet} is one of the first attempts to use Deep Learning for semantic segmentation on point set data in a more efficient way. While PointNet could capture the global structure of a point set by pooling over individual point features, it could not capture local structure. The authors bridged that gap with PointNet++ \cite{pointnet-plus} in which instead of global pooling, the network learns descriptors of neighborhoods in the point set. By varying the size of these neighborhoods, the network is able to learn a hierarchical description of the underlying shape, similar to how a Convolutional Neural Network learns the representation of a 2D image or 3D voxel volume.

\section{Technical Approach}\label{sec:technical_approach}
We first explain the reason for SGPN's memory intensiveness and then discus the two approaches we experimented with. For a Point Set with $N$ points, SGPN first uses the PointNet networks to compute a feature matrix $F_{sim}$ of size $N \times N_f$ where $N_f$ is a hyperparameter representing size of one point's feature representation. SGPN then computes an $N \times N$ Similarity matrix $S$ whose $i,j^{th}$ entry is given by the Euclidean distance between the feature representations of the $i,j^{th}$ points, that is, $S_{ij} = \lVert F_{sim}^i - F_{sim}^j \rVert_2$. Each row $i$ of $S$ can be understood as Group Proposal for the $i^{th}$ point, i.e. if points $i$ and $j$ lie in the same instance then we want the value of $S_{ij}$ to be small. In this manner we are forcing the feature representations produced by PointNet networks to be closer to each other in the feature space, even if they are distant in the Euclidean space.

While during Training, the authors fix the number of points they use for a Point Set, this Similarity Matrix whose size is quadratic in the number of points, can blow up during Test time, if we have a scene with a large number of points. We propose two sub-sampling based approaches to resolve this issue - the general idea is to compute Instance Segmentation on a subset of points of a scene, and then propagate their predicted labels to all points.

\begin{figure}[t]
\centering
 \includegraphics[width=1.0\linewidth]{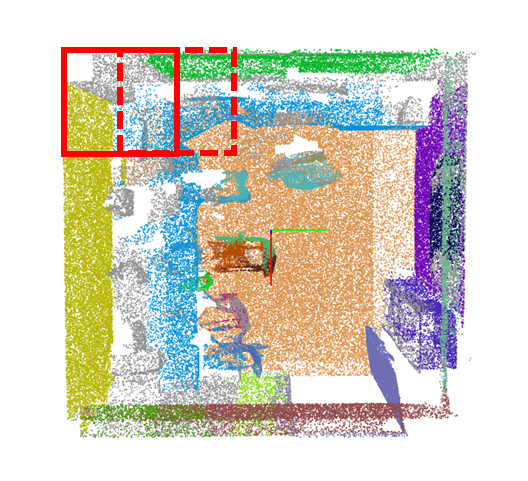}
\caption{Overlapping Blocks shown for one of the scenes (This is top view of the scene).}
\label{fig:blocks}
\end{figure}

\textbf{Block Processing}: The way SGPN works is that it divides a scene into $1m \times 1m$ cuboidal blocks with overlaps of stride $0.5m$ (the height of each block is the same as the height of the scene). Figure \ref{fig:blocks} shows an example of a scene with two such overlapping blocks in red. With SGPN, while training, the authors use random sub-sampling to ensure that each block has $4096$ points in it. During test time, they use the actual number of points $N$ to create a $N \times N$ similarity matrix, which is why this memory issue only appears at test time. SGPN processes each block individually to predict Instance Labels, and at the end these block labels are merged using an algorithm proposed by the authors. For the rest of the paper, we'll treat already pre-processed scenes which have $4096$ points in each block. Our approach can be generalized and give labels to any number of points in the scene, including those with too many points for a similarity matrix to be computed.

\subsection{Random Sub-sampling}
In this approach we randomly select $K$ points out of 4096 in every block in a scene, and use a pre-trained SGPN network to find an Instance Segmentation Label for each of the $K$ points. We then propagate these labels to all the 4096 points. Specifically, for each point $p$ in the scene, we find the point $q_K(p)$ among the $K$ points which is nearest in terms of Euclidean distance to $p$. We then set $\textbf{Label}(p) = \textbf{Label}(q_K(p))$. To find the nearest point in the last step, we use an implementation of KDTrees provided by the SciPy Python package. We construct a KDTree with the $K$ randomly chosen points once - this takes $O(K\log K)$ and thereafter each search on average taken $O(\log K)$ time. Therefore if we do $N$ searches, it should take on average $O(N\log K)$ time.

\subsection{Grid-based Sub-sampling}
\begin{figure}[t]
\centering
 \includegraphics[width=1.0\linewidth]{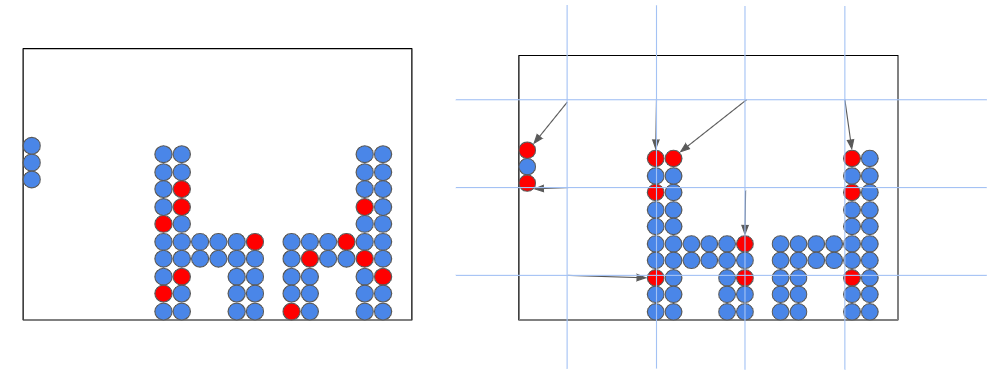}
\caption{Random sub-sampling method on the left misses an entire object on the left as it focuses on high density parts of the scene whereas these points would be captured by the grid-based sub-sampling method as shown on the right.}
\label{fig:grid-justification}
\end{figure}

Random sub-sampling is just one way of choosing landmark points to do instance segmentation on before propagating the instance information to the respective neighbours. However, it has its limits as we have no control at all over which points are selected using this method. A more controlled approach we came up with uses a regular grid to evenly divide the scene block into small cubes. The vertices of these small cubes are evenly spaced around the scene block. We still want control over the number of landmarks that are chosen by our algorithm. We thus take the n-closest scene block neighbours of these grid points as a sub-sampling of the scene to get these K landmarks. A more detailed algorithm description of this sub-sampling approach can be found in Algorithm \ref{grid-algo}. As illustrated in figure \ref{fig:grid-justification}, our hope is that this sampling method would allow us to capture patterns that would have been missed by the random sub-sampling method.

\begin{algorithm}
  \begin{algorithmic}
    \STATE{\textit{Input}}
    \begin{itemize}
        \item[--] \textbf{K}: number of desired landmarks
        \item[--] \textbf{block}: 3D coordinates of all points in the scene block
        \item[--] \textbf{nmin}, \textbf{nstep}: speed-up parameters, usually both are set to 1
        \item[--] \textbf{grid}: evenly-spaced point grid over the space of the scene block
    \end{itemize}
    \STATE{\textit{Output}}
    \begin{itemize}
        \item[--] \textbf{landmarks}: K landmark points chosen through the grid-based approach
    \end{itemize}
    \STATE{\textit{Algorithm}}
    \STATE nearest\_idx = [], n = nmin
    \WHILE{len(nearest\_idx) < K}
    \STATE nearest\_idx = []
    \FOR{grid\_point in grid}
        \STATE Append to nearest\_idx the indices of the n closest element of the scene block to grid\_point
    \ENDFOR
    \STATE n += nstep
    \ENDWHILE
    \STATE landmarks = randomly chosen K samples of nearest\_idx without replacement
  \end{algorithmic}
  \caption{Grid-based sub-sampling}
  \label{grid-algo}
\end{algorithm}

\begin{figure}[t]
\centering
 \includegraphics[width=1.0\linewidth]{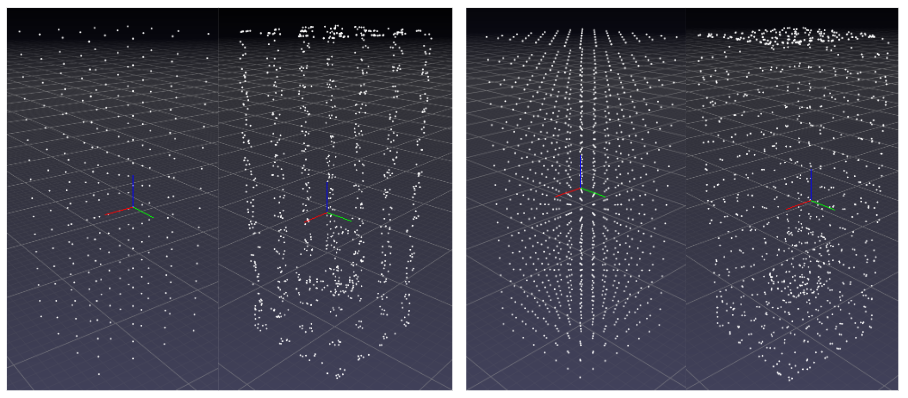}
\caption{Two density configurations of the grid with the corresponding chosen landmarks for the same scene. On the left, the n-neighbours create small clustering patterns due to the small density of the grid in comparison with the right configuration.}
\label{fig:grid_density}
\end{figure}

The number of points chosen to create the grid (or equivalently its density) is a hyperparameter of our algorithm. As shown in figure \ref{fig:grid_density}, a poor choice of density may create unwanted patterns in the landmarks chosen by our algorithm. For the scenes we treated, 2048 points evenly spaced in the grid seemed to be dense enough to get rid of these patterns and to produce good results in a reasonable amount of time.

\begin{figure}[t]
\centering
 \includegraphics[width=1.0\linewidth]{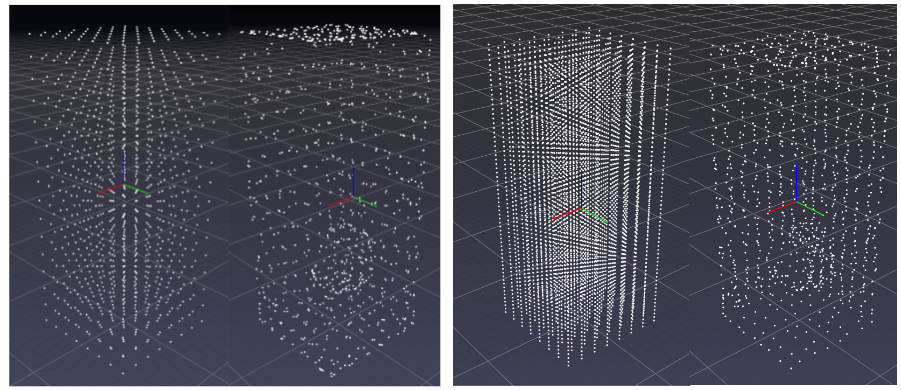}
\caption{Finding $K=1024$ landmarks for the same scene block using algorithm \ref{grid-algo} on the left and algorithm \ref{gridext-algo} on the right. A grid size of $1024$ was sufficient to produce the results on the left whereas $4096$ points were needed for the same-quality result on the right.}
\label{fig:gridext-vs-grid}
\end{figure}

Instead of using n-neighbours that can create cluster patterns as shown in figure \ref{fig:grid_density}, we tried adjusting the density of the grid to each scene block in order to capture the $K$ landmarks with only a 1-nearest-neighbour approach. The details are shown in algorithm \ref{gridext-algo}. However, with this approach, the grid density needed for the same number of landmarks is increasingly more important as shown in figure \ref{fig:gridext-vs-grid}. We have less control on the memory that we use, which tends to blow up with this strategy. That is why we've decided to keep algorithm \ref{grid-algo} to produce our results after carefully choosing our grid size to prevent unwanted clustering patterns.

\begin{algorithm}
  \begin{algorithmic}
    \STATE{\textit{Input}}
    \begin{itemize}
        \item[--] \textbf{K}: number of desired landmarks
        \item[--] \textbf{block}: 3D coordinates of all points in the scene block
    \end{itemize}
    \STATE{\textit{Output}}
    \begin{itemize}
        \item[--] \textbf{landmarks}: K landmark points chosen through the gridextension-based approach
    \end{itemize}
    \STATE{\textit{Algorithm}}
    \STATE nearest\_idx = [], n\_grid = 128
    \WHILE{len(nearest\_idx) < K}
    \STATE nearest\_idx = []
    \STATE grid = create an evenly spaced grid of size n\_grid
    \FOR{grid\_point in grid}
        \STATE Append to nearest\_idx the index of the closest element of the scene block to grid\_point
    \ENDFOR
    \STATE n\_grid = 2 * n\_grid
    \ENDWHILE
    \STATE landmarks = randomly chosen K samples of nearest\_idx without replacement
  \end{algorithmic}
  \caption{GridExtension-based sub-sampling}
  \label{gridext-algo}
\end{algorithm}

\section{Results}
Figure \ref{fig:teaser} shows the results of our sub-sampling based Instance Segmentation methods for one scene from the Stanford Indoor 3D Dataset \cite{stanford-dataset}. We can observe for both sub-sampling approaches that as we increase the number of points in our sample, the quality of instance segmentation seems to improve.

\subsection{Random Sub-sampling}
\begin{figure}[t]
\centering
 \includegraphics[width=1.0\linewidth]{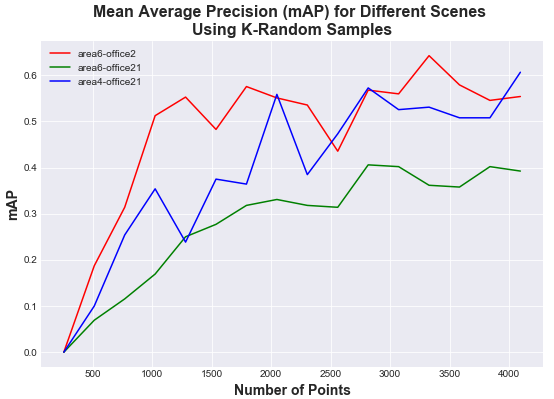}
\caption{\textbf{Random Sub-sampling}: Mean Average Precision for Instance Segmentation for three scenes from the Stanford Indoor 3D Dataset \cite{stanford-dataset}. }
\label{fig:random-plot}
\end{figure}

\begin{figure}[t]
\centering
 \includegraphics[width=1.0\linewidth]{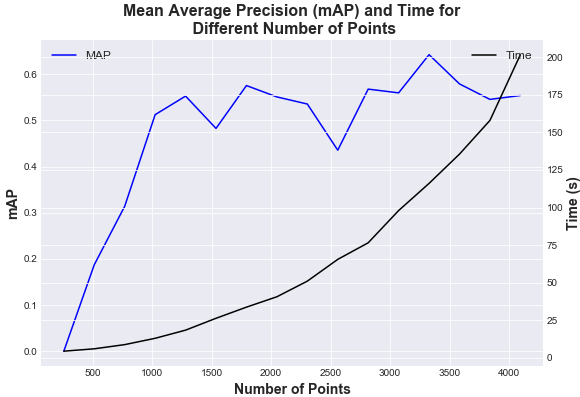}
\caption{\textbf{Random Sub-sampling}: Mean Average Precision and Computation Time for the scene "Area 6, Office 2" from the Stanford Indoor 3D Dataset \cite{stanford-dataset}. }
\label{fig:random-time}
\end{figure}

Figure \ref{fig:random-plot} plots the Mean Average Precision (mAP) \cite{sgpn} for Instance Segmentation and the Computation Time for three scenes from the Stanford Indoor 3D Dataset. We can observe that using half the points (i.e. $2048$ out of $4096$) on average, the performance as measured by mAP is only slightly worse compared to the performance with using the full Point Set (of size $4096$), albeit with only $1/4^{th}$ the original memory consumption (as the size of the Similarity Matrix is quadratic in number of points).  Moreover, Figure \ref{fig:random-time} shows that when we use only $2048$ points, the computation time is only $62s$ which is less than a third of the original time ($201s$).

\begin{figure}[t]
\centering
 \includegraphics[width=1.0\linewidth]{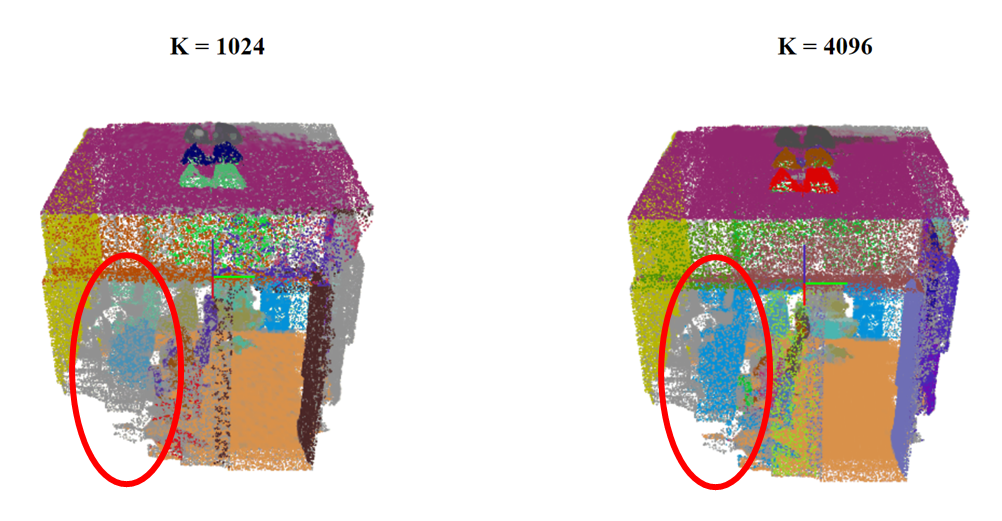}
\caption{Predicted Instance Segmentation using the full Point Set (4096 points) and a Random Sub-sample of 1024 points for the scene "Area 6, Office 2" from the Stanford Indoor 3D Dataset \cite{stanford-dataset}. The red oval area shows some deficiencies which crop in due to sub-sampling.}
\label{fig:random-tradeoff}
\end{figure}

We take a closer look at the Instance Segmentation generated using sub-sampling in Figure \ref{fig:random-tradeoff}. We can observe that using full $K = 4096$ points, SGPN was able to correctly assign the label table to all points inside the table (the blue area inside the red oval). However, using a sub-sample of size $1024$, some of the points inside the table (see the red oval) were incorrectly classified as belonging to the wall - these points are marked in grey color. This is not a surprise because if we have two points in our random sample, call them $p_{left}$ and $p_{right}$, and there is a plane which is equidistant from the two points and perpendicular to the line joining them, then all points to the left of this plane will be assigned the label $\text{Label}(p_{left})$ and all points to the right of this plane will be assigned $\text{Label}(p_{right})$.

\subsection{Grid Sub-sampling}

\begin{figure}[t]
\centering
 \includegraphics[width=1.0\linewidth]{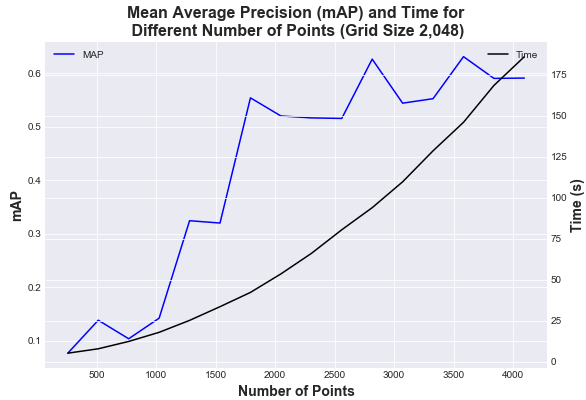}
\caption{\textbf{Grid Sub-sampling}: Mean Average Precision and Computation Time for the scene "Area 6, Office 2" from the Stanford Indoor 3D Dataset \cite{stanford-dataset}. }
\label{fig:random-grid}
\end{figure}

As shown in figure \ref{fig:random-grid}, our grid sub-sampling method starts giving reasonably good results at around 1700 landmarks chosen. We can clearly see the before and after difference in the bottom row of figure \ref{fig:teaser}.

One thing that we can notice on figure \ref{fig:grid-justification} is that even if the mAP score can be lower with a smaller number of landmarks, the output can  be smoothed out and turn out better in certain regions of the scene. In this particular example, the red-circled chair has a better instance segmentation for $K=2048$ in comparison with the full $K=4096$ points in the scene. 

\begin{figure}[t]
\centering
 \includegraphics[width=1.0\linewidth]{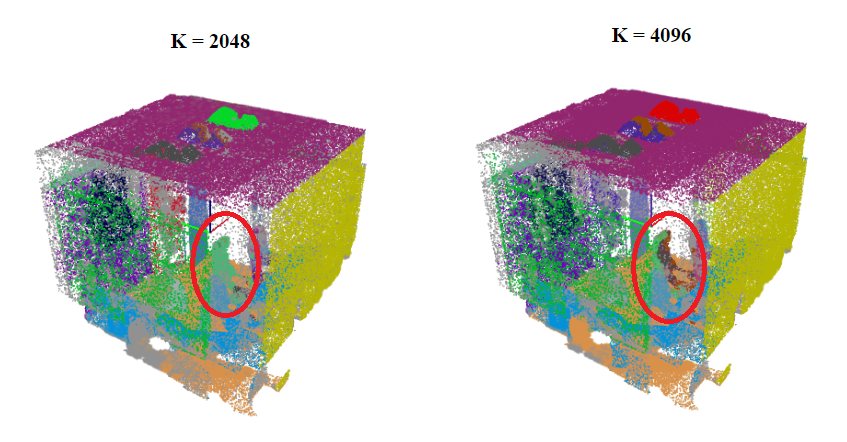}
\caption{Comparison of instance segmentations for two different values of the number of wanted landmark points. The red-cirled chair has a better instance segmentation on the left despite a smaller mAP score as can be seen in figure \ref{fig:random-grid}}.
\label{fig:grid_chair_comparison}
\end{figure}

\section{Discussion}
\begin{figure}[t]
\centering
 \includegraphics[width=1.0\linewidth]{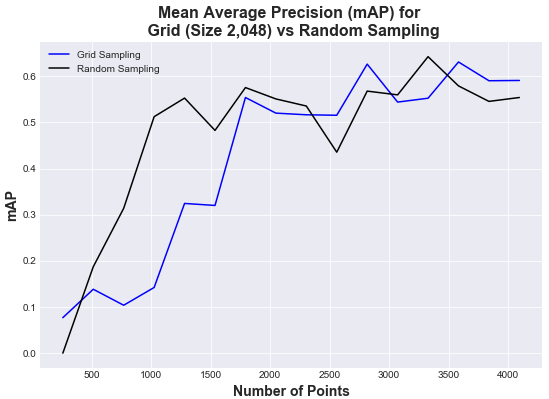}
\caption{Mean Average Precision comparison for both approaches for the scene "Area 6, Office 2" from the Stanford Indoor 3D Dataset \cite{stanford-dataset}.}
\label{fig:map-comparison}
\end{figure}

Both approaches give similar results for higher number of desired landmarks. However, the random sub-sampling method gives better results with a smaller number of landmarks chosen and can thus help decrease more memory consumption and running time. This difference can be seen with the visuals of figure \ref{fig:teaser} or more quantitatively with the mAP comparison of figure \ref{fig:map-comparison}. This difference is due to the density of the grid we're using to compute the landmarks. Having the number of points of the grid go to infinity, we end up with the same algorithm as the random sub-sampling one due to the sampling to get the landmark points from algorithm \ref{grid-algo}. However, we want a less dense grid to tackle the issue of figure \ref{fig:grid-justification}, and to have reasonable time and space complexities for algorithm \ref{grid-algo}.

\section{Conclusion}
Saving memory and time are challenges that need to be overcome to efficiently use Deep Learning approaches with Point Sets. SGPN which does instance segmentation by using PointNet++ was a big step in this direction. However, it required the usage of a dense matrix whose shape was quadratic in the number of points in the scene block. We managed to reduce this shape to a fixed $K \times K$ matrix while significantly accelerating the production of the output. While both approaches seem to perform equally well on a high number of landmarks, it seems that random sub-sampling should be preferred to save even more memory for particularly challenging scenes. Grid-based sub-sampling could still be useful if the problem shown in figure \ref{fig:grid-justification} is particularly harmful to the segmentation of a scene. Among the future work that could be done, there is the possibility to come up with better strategies to get landmark points. Another crucial point we have not mentioned a lot in this paper is the different strategies we can use to propagate the information from landmarks to other points. One specific thing that could be tried is propagate the instance to neighbours in the feature space with the Euclidian distance. In that space, points which are close together should come from the same instance, as the SGPN network was optimized on this task. This requires the usage of the $N\times N_{features}$ matrix. The rest of the network can be discarded at test time so that we do not need to build the unwanted $N \times N$ matrix. We would still need the $K \times K$ matrix of the landmark points to perform instance segmentation.

\section{Contributions}

While Abhi focused more on the random sub-sampling method and Julien on the grid-based approach, we pair-coded most of the implementations together and have equal contributions for this project.

\bibliographystyle{eg-alpha-doi}
\bibliography{6838bibsample}

\newcommand{\etalchar}[1]{$^{#1}$}
\begin{thebibliography}{\uppercase{GOO{\etalchar{*}}17b}}

\bibitem[ASRZ{\etalchar{*}}16]{stanford-dataset}
\textsc{Armeni I., Sener O., R.~Zamir A., Jiang H., Brilakis I., Fischer M.,
  Savarese S.}:
\newblock 3d semantic parsing of large-scale indoor spaces.
\newblock pp.~1534--1543.
\newblock \href {http://dx.doi.org/10.1109/CVPR.2016.170}
  {\path{doi:10.1109/CVPR.2016.170}}.

\bibitem[GOO{\etalchar{*}}17a]{voxnet}
\textsc{Garcia{-}Garcia A., Orts{-}Escolano S., Oprea S., Villena{-}Martinez
  V., Rodr{\'{\i}}guez J.~G.}:
\newblock A review on deep learning techniques applied to semantic
  segmentation.
\newblock \emph{CoRR abs/1704.06857} (2017).
\newblock URL: \url{http://arxiv.org/abs/1704.06857}, \href
  {http://arxiv.org/abs/1704.06857} {\path{arXiv:1704.06857}}.

\bibitem[GOO{\etalchar{*}}17b]{segmentation}
\textsc{Garcia{-}Garcia A., Orts{-}Escolano S., Oprea S., Villena{-}Martinez
  V., Rodr{\'{\i}}guez J.~G.}:
\newblock A review on deep learning techniques applied to semantic
  segmentation.
\newblock \emph{CoRR abs/1704.06857} (2017).
\newblock URL: \url{http://arxiv.org/abs/1704.06857}, \href
  {http://arxiv.org/abs/1704.06857} {\path{arXiv:1704.06857}}.

\bibitem[LB98]{cnn}
\textsc{LeCun Y., Bengio Y.}:
\newblock The handbook of brain theory and neural networks.
\newblock MIT Press, Cambridge, MA, USA, 1998, ch.~Convolutional Networks for
  Images, Speech, and Time Series, pp.~255--258.
\newblock URL: \url{http://dl.acm.org/citation.cfm?id=303568.303704}.

\bibitem[LS18]{megadepth}
\textsc{Li Z., Snavely N.}:
\newblock Megadepth: Learning single-view depth prediction from internet
  photos.
\newblock \emph{CoRR abs/1804.00607} (2018).
\newblock URL: \url{http://arxiv.org/abs/1804.00607}, \href
  {http://arxiv.org/abs/1804.00607} {\path{arXiv:1804.00607}}.

\bibitem[QSMG16]{origpointnet}
\textsc{Qi C.~R., Su H., Mo K., Guibas L.~J.}:
\newblock Pointnet: Deep learning on point sets for 3d classification and
  segmentation.
\newblock \emph{CoRR abs/1612.00593} (2016).
\newblock URL: \url{http://arxiv.org/abs/1612.00593}, \href
  {http://arxiv.org/abs/1612.00593} {\path{arXiv:1612.00593}}.

\bibitem[QYSG17]{pointnet-plus}
\textsc{Qi C.~R., Yi L., Su H., Guibas L.~J.}:
\newblock Pointnet++: Deep hierarchical feature learning on point sets in a
  metric space.
\newblock In \emph{Advances in Neural Information Processing Systems 30}, Guyon
  I., Luxburg U.~V., Bengio S., Wallach H., Fergus R., Vishwanathan S., Garnett
  R., (Eds.). Curran Associates, Inc., 2017, pp.~5099--5108.
\newblock URL:
  \url{http://papers.nips.cc/paper/7095-pointnet-deep-hierarchical-feature-learning-on-point-sets-in-a-metric-space.pdf}.

\bibitem[WYHN17]{sgpn}
\textsc{Wang W., Yu R., Huang Q., Neumann U.}:
\newblock {SGPN:} similarity group proposal network for 3d point cloud instance
  segmentation.
\newblock \emph{CoRR abs/1711.08588} (2017).
\newblock URL: \url{http://arxiv.org/abs/1711.08588}, \href
  {http://arxiv.org/abs/1711.08588} {\path{arXiv:1711.08588}}.

\bibitem[ZPIE17]{CycleGAN2017}
\textsc{Zhu J.-Y., Park T., Isola P., Efros A.~A.}:
\newblock Unpaired image-to-image translation using cycle-consistent
  adversarial networks.
\newblock In \emph{Computer Vision (ICCV), 2017 IEEE International Conference
  on} (2017).

\end{thebibliography}

\end{document}